%% file: RSS2016-rearrangement.tex
\documentclass[conference]{IEEEtran}
\usepackage[sc]{mathpazo}
\usepackage[T1]{fontenc}
\usepackage{hyperref}
\hypersetup{bookmarksopen,bookmarksnumbered,
pdfpagemode=UseOutlines,
colorlinks=true,
linkcolor=blue,
anchorcolor=blue,
citecolor=blue,
filecolor=blue,
menucolor=blue,
urlcolor=blue
}

\usepackage[numbers,sort&compress]{natbib}
\usepackage{multicol}

\usepackage{algorithm}
\usepackage[noend]{algpseudocode}
\usepackage[cmex10]{amsmath}
\usepackage{amssymb}
\usepackage{array} 
\usepackage{booktabs}
\usepackage[font=footnotesize]{caption}
\usepackage{color}
\usepackage{ifthen,version}
\usepackage{import}
\usepackage[font=footnotesize]{subcaption}
\usepackage{url}
\usepackage{verbatim}
\usepackage{flushend}
\usepackage[pdftex]{graphicx}
\usepackage{siunitx}
\DeclareGraphicsExtensions{.pdf,.jpeg,.jpg,.png}

\newcommand{\eref}[1]{Eq.~\ref{#1}}
\newcommand{\sref}[1]{Section~\ref{#1}}
\newcommand{\figref}[1]{Fig.\ref{#1}}
\newcommand{\aref}[1]{Algorithm~\ref{#1}}

\newenvironment{proof}[1][Proof]{\begin{trivlist}
\item[\hskip \labelsep {\bfseries #1}]}{\end{trivlist}}
\newenvironment{proposition}[1][Proposition]{\begin{trivlist}
\item[\hskip \labelsep {\bfseries #1}]}{\end{trivlist}}

\newcommand{\Robot}{\mathcal{R}}
\newcommand{\Movables}{\mathcal{M}}

\newcommand{\Obstacles}{\mathcal{O}}
\newcommand{\Cspace}{\mathcal{C}}
\newcommand{\Crobot}{\Cspace^R}
\newcommand{\X}{\mathcal{X}}
\newcommand{\Xfree}{\mathcal{X}_{free}}
\newcommand{\Xgoal}{\mathcal{X}_G}
\newcommand{\goalobj}{g}
\newcommand{\Uspace}{\mathcal{U}}

\DeclareMathOperator{\dist}{d}
\DeclareMathOperator{\fk}{FK}
\DeclareMathOperator{\cost}{c}

\newcommand{\chatcontact}{\hat{\dist}_{con}}

\newcommand{\chatmove}{\hat{\dist}_{move}}

\newcommand{\Basic}{\mathcal{B}}
\newcommand{\Queue}{\mathcal{Q}}

\newcommand{\Vertices}{V}

\newcommand{\setwo}{SE(2)}

\newcommand{\assumptionee}{Assumption 1}
\newcommand{\assumptionchain}{Assumption 2}
\newcommand{\assumptionquasi}{Assumption 3}

\newcommand{\hypothesis}[2]{\textbf{#1}{#2}}

\newcommand{\pushingrrt}{\textbf{\textit{Pushing RRT}}~}
\newcommand{\primitiverrt}{\textbf{\textit{Primitive RRT}}~}
\newcommand{\basicsearch}{\textbf{\textit{Basic Search}}~}
\newcommand{\pushingsearch}{\textbf{\textit{Pushing Search}}~}

\newsavebox{\largestimage}

\newboolean{include-notes}
\setboolean{include-notes}{true}
\newcommand{\ssnote}[1]{\ifthenelse{\boolean{include-notes}}%
  {\textcolor{blue}{\textbf{SS: #1}}}{}}
\newcommand{\jknote}[1]{\ifthenelse{\boolean{include-notes}}%
 {\textcolor{green}{\textbf{JK: #1}}}{}}
\newcommand{\mcnote}[1]{\ifthenelse{\boolean{include-notes}}%
 {\textcolor{orange}{\textbf{MC: #1}}}{}}

\renewcommand{\baselinestretch}{1.04}

\begin{document}

\title{Rearrangement Planning via Heuristic Search}

\author{
    \IEEEauthorblockN{Jennifer E.~King, Siddhartha S.~Srinivasa}
    \IEEEauthorblockA{Carnegie Mellon University (CMU)}
    \IEEEauthorblockA{
      \{jeking, ss5\}@andrew.cmu.edu}
  }


%

\maketitle

\import{.}{abstract.tex}

\IEEEpeerreviewmaketitle

\import{.}{introduction.tex}

\import{.}{problem_definition.tex}
\import{.}{search_based.tex}

\import{.}{results.tex}

\import{.}{discussion.tex}

\import{.}{acknowledgments.tex}


\bibliographystyle{plainnat}
\bibliography{references}

\end{document}

%% file: abstract.tex
\begin{abstract}
We present a method to apply heuristic search algorithms to solve
\textit{rearrangement planning by pushing} problems. In these
problems, a robot must push an object through clutter to achieve a
goal. To do this, we exploit the fact that contact with objects in the
environment is critical to goal achievement. We dynamically generate
goal-directed primitives that \textit{create} and \textit{maintain}
contact between robot and object at each state expansion during the
search. These primitives focus exploration toward critical areas of
state-space, providing tractability to the high-dimensional planning
problem. We demonstrate that the use of these primitives, combined
with an informative yet simple to compute heuristic, improves success
rate when compared to a planner that uses only primitives formed from
discretizing the robot's action space. In addition, we show our
planner outperforms RRT-based approaches by producing shorter paths
faster.  We demonstrate our algorithm both in simulation and on a
7-DOF arm pushing objects on a table.
\end{abstract}

%% file: introduction.tex

\section{Introduction}
\label{sec:introduction}

\begin{figure}
\includegraphics[width=\columnwidth]{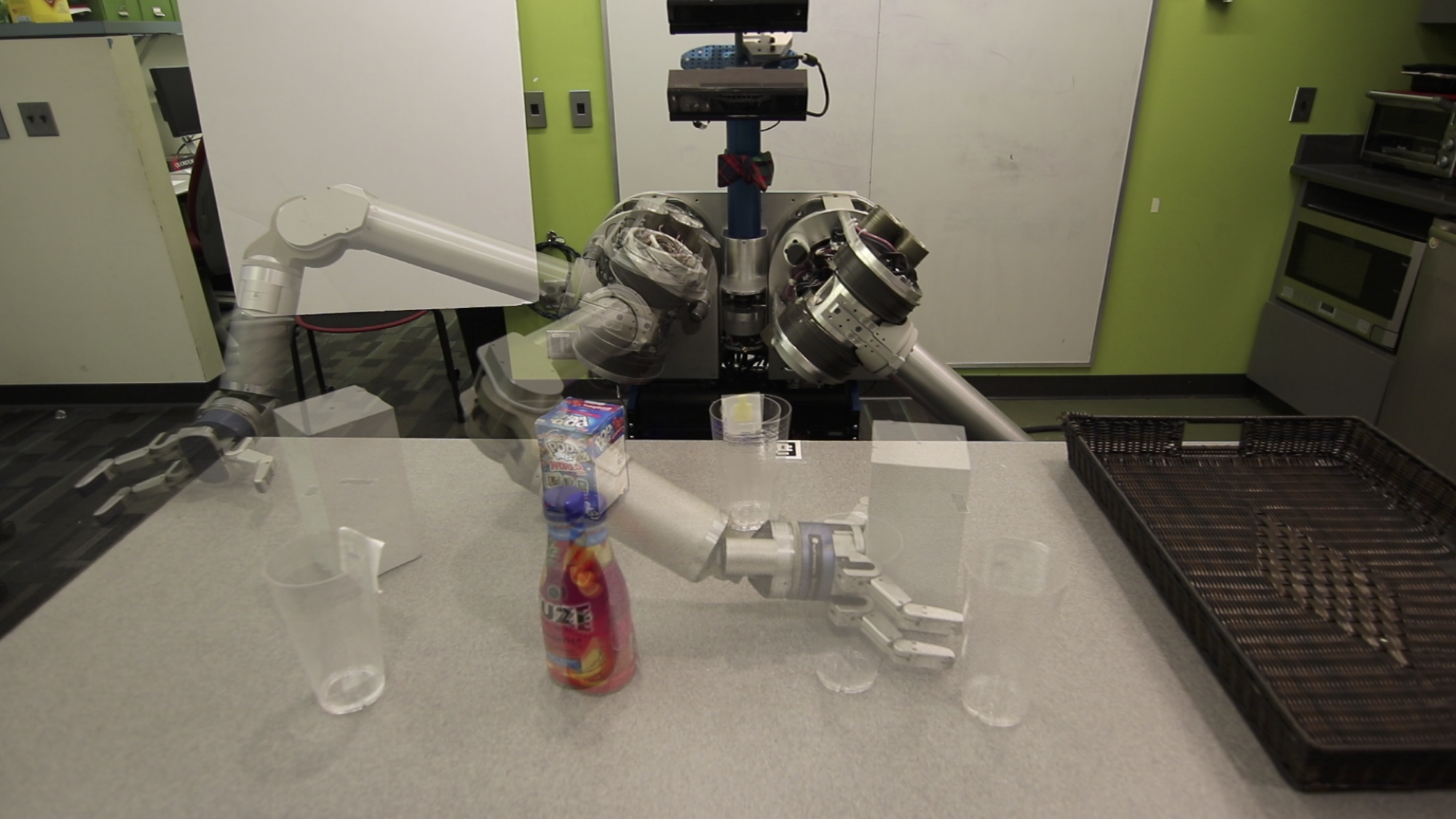}\\
\caption{A solution to a rearrangement problem. The robot moves the box across the table to
  make it reachable by the left arm. The robot freely interacts with
  and moves clutter in the scene in order to achieve the goal.}
\label{fig:figure1}
\end{figure}

In this paper we present a method for planning
pushing actions that allow a robot to move an object to a goal
pose through clutter. Consider the scene
in~\figref{fig:figure1}. Here the robot is tasked with moving the white block
from the right side of the table to the left side of the table in
order to make it accessible to the left arm to lift and place the block
on the tray. All objects on the table can be moved by the robot and
the final pose of these objects does not matter.

Rearrangement
planning~\cite{dogar11pushgrasping,nieuwenhuisen08namo,stilman04namo,vandenburg08planning,barry12darrt,stilman07namo}
problems such as this are difficult for two reasons. First, the
planner must search the high-dimensional state space containing the
state of the robot and all the objects cluttering the scene.

Second, the system is underactuated and non-holonomically
constrained. In particular, the objects in the environment can only
move when contacted by the robot, and the motion of a contacted object
is directly governed by the physics of the contact.

As a consequence, solving the two-point boundary value problem (2PTBVP) ---
connecting two states via feasible control inputs --- is often analytically 
intractable and numerically expensive. This is in contrast with holonomically
constrained systems, where often straight lines (more technically, geodesics
in configuration space) trivially solve the 2PTBVP.

Randomized planning methods such as the Kinodynamic Rapidly-Exploring Random Tree
(RRTs)~\cite{lavalle98rrt} have been shown to be quite effective at
handling high-dimensional state spaces and have proven applicable to
the rearrangement problem~\cite{haustein15kinodynamic,
  king15nonprehensile, zickler09physics} by never needing to solve the 2PTBVP
  exactly, instead sampling and rolling out actions and growing trees that 
  never have to meet at a point. However, the resulting
solutions are often quite suboptimal. Post-processing methods such as
shortcutting~\cite{hauser10shortcut, sanchez02collisionchecking,
  sekhavat98multilevel} can be applied to improve the local optimality
of solutions, but these solutions may still be far from the global
optimal.

Proposed variants to the RRT algorithm allow the planner to find
globally optimal solutions over time~\cite{karaman11optmotplan,
  webb12kinorrt, gammell14irrt, gammell15bit}. But these methods
depend on the ability to solve the 2PTBVP in order to ``rewire''
suboptimal paths through the graph.

\begin{figure*}[ht!]
\captionsetup[subfigure]{margin=5pt}
\begin{subfigure}[b]{0.2\linewidth}
\centering
\includegraphics[width=0.98\linewidth]{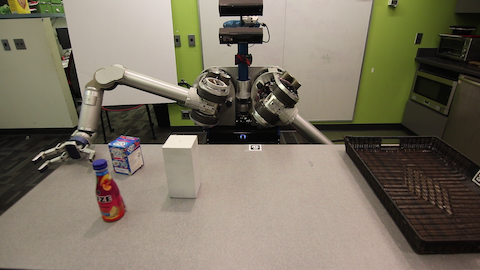}
\end{subfigure}%
\begin{subfigure}[b]{0.2\linewidth}
\centering
\includegraphics[width=0.98\linewidth]{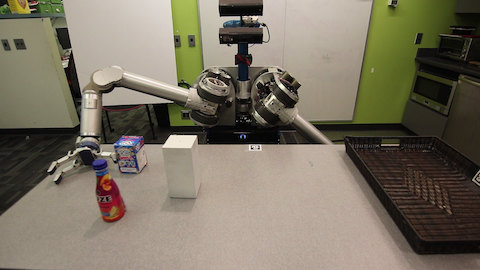}
\end{subfigure}%
\begin{subfigure}[b]{0.2\linewidth}
\centering
\includegraphics[width=0.98\linewidth]{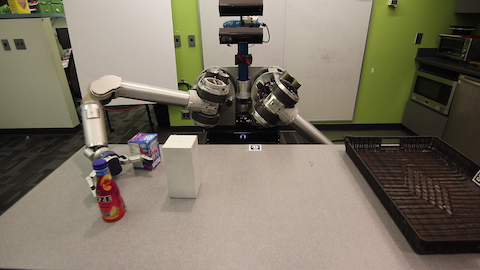}
\end{subfigure}%
\begin{subfigure}[b]{0.2\linewidth}
\centering
\includegraphics[width=0.98\linewidth]{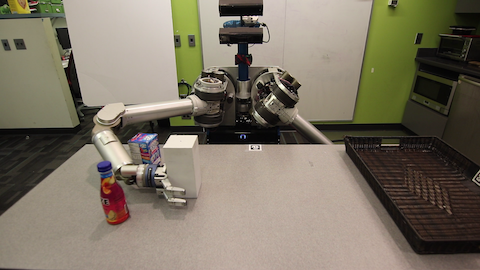}
\end{subfigure}%
\begin{subfigure}[b]{0.2\linewidth}
\centering
\includegraphics[width=0.98\linewidth]{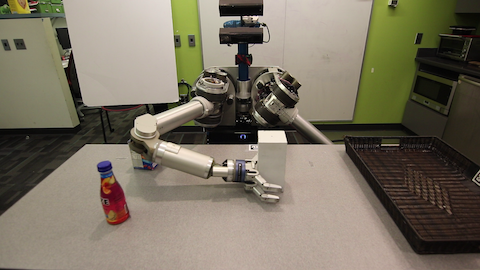}
\end{subfigure}
\caption{A rearrangement planning problem.  The robot is tasked with
  pushing the white box near the tray. The plan exhibits
  \textit{contact} and \textit{pushing} primitives to
  move clutter and achieve the goal.}
\label{fig:filmstrip_scene10}
\end{figure*}

Likewise, graph-based methods
such as A* search~\cite{hart68astar} are attractive
because they find globally optimal solutions when applied to discrete
state and action spaces. Prior works have shown their applicability to
continuous state and action spaces by first discretizing the state
space, then connecting the discrete states with feasible
actions~\cite{chen98sandros, likhachev08lattice,
  cohen10motionprimitives}. However, even these methods require a solution
  to the 2PTBVP for creating the lattice.

Our first key insight is that, although we cannot solve the 2PTBVP in the full
state space, we can solve it in the lower dimensional subspace
containing only the robot. In other words, we can generate actions that move the
robot between two configurations. This allows us to generate
primitives by specifying desired configurations for the robot.

  However, we are faced with two more challenges:
\begin{enumerate}
\item \textbf{Action space:} Typical lattice methods define a single
  set of actions, or primitives, to be applied at every state. Contact
  between robot and objects is critical to success in rearrangement
  planning and the set of primitives that create contact between robot
  and objects varies with state. This makes it difficult to define a
  single set of primitives that is useful across all states.

\item \textbf{Graph resolution:} Discretizing the continuous state
  space allows the planning algorithm to consider nearby states
  equivalent.  This relies on the assumption that two nearby states
  will remain nearby when a primitive is applied. The discontinuous
  nature of pushing interaction means this assumption often does not
  hold. Even a very small change in the initial pose of an object can
  lead to very different final poses of the object after being
  pushed. Thus, applying naive discretization methods can lead us to
  incorrectly define two states equivalent, causing the search to miss
  important areas of state space.
\end{enumerate}

Our second key insight is that we can dynamically generate primitives that
create contact with objects at each state expansion during the
search. We can use these dynamically generated primitives to supplement a
core primitive set applied to every state. This focuses exploration
toward critical areas of state space.  The focused search, in turn,
allows us to use a very fine graph resolution.

To dynamically generate primitives we exploit two further aspects of
the problem: (1) we must \textit{create} contact and
(2) in the quasistatic environments we consider, we must
\textit{maintain} contact in order to move an object.  We define
\textit{contact} primitives as those that move the robot to a
configuration in or near contact with an object.  Then, we use simple
physics assumptions to generate \textit{pushing} primitives that move
the robot in a direction likely to maintain contact with an object.


The remainder of this paper is organized as
follows. In~\sref{sec:problem_definition} we define the rearrangement
planning problem. We then define a core set of \textit{basic}
primitives, and the dynamically generated \textit{contact} and
\textit{pushing} primitives in~\sref{sec:search_based}. In addition,
we show how these primitives, combined with an informative yet simple
to compute heuristic, can be used in discrete search. We present
experimental results in~\sref{sec:results} that demonstrate that the
use of \textit{contact} and \textit{pushing} primitives improves
success rate of a search based planner when compared to one that uses
only \textit{basic} primitives. In addition, we show that our planner
produces shorter paths faster when compared to RRT-based approaches
for solving the same problem.

%% file: problem_definition.tex
\section{The Rearrangement Planning Problem}
\label{sec:problem_definition}

Assume we have a robot, $\Robot$, working in a bounded world populated
with a set of objects $\Movables$ that the robot is allowed to
manipulate and a set of obstacles $\Obstacles$ which the robot is
forbidden to contact. The robot is endowed with configuration space
$\Crobot$ and each object in $\Movables$ is endowed with configuration
space $\Cspace^i$ for $i = 1 \dots n$.

Our search is defined on the state-space $\X$ that is the Cartesian
product space of the configuration spaces of the robot and the objects
in $\Movables$: $\X = \Cspace^R \times \Cspace^1 \times \dots
\times \Cspace^m$. A state $x \in \X$ is defined by $x = \left(q, o^1,
\dots, o^m\right), q \in \Cspace^R, o^i \in \Cspace^i$ $\forall i$. We
define the free state space $\Xfree \subseteq \X$ as the set of all
states where the robot and objects are not penetrating themselves, the
obstacles or each other. We allow contact between robot and objects in
all states in $\Xfree$.  This contact is critical for manipulation.

The state $x$ evolves nonlinearly
based on the physics of the manipulation. The motion of the
objects is governed by the contact between the objects and the
manipulator.  We describe this evolution as a non-holonomic
constraint:
\begin{equation}
\dot{x} = f(x, u)
\label{eq:nonholonomic_constraint}
\end{equation}
where $u \in \Uspace$ is an instantaneous control input.  The function
$f$ encodes the physics of the environment. 

We define a primitive, $a$, as a discrete set of control inputs:
\begin{equation}
a = \{(u, d)_i | u \in \Uspace, d \in \mathbb{R}^+, i=1 \dots p\}
\end{equation}
where $u$ defines a control and $d$ defines the duration to apply the
control.

We identify a single object, $\goalobj \in \Movables$ as the
\textit{goal object}. We identify a goal region $\Xgoal$ as the set of
states where where the goal object is within radius $r_{goal}$ of a
desired configuration $p_{goal} \in C^g$:
\begin{equation}
\Xgoal = \{ x \vert x \in \Xfree, o^g \in x, \|o^g - p_{goal}\| \leq r_{goal} \}
\end{equation}

The task of rearrangement planning is to find a sequence of primitives (path), $\pi: =
\{a_1 \dots a_t\}$, such that when applied to a start state $x_0 \in
\Xfree$ under the constraint $f$ we end in the goal region.

We define the cost of a path, $\pi$, as the
distance the end-effector of the robot moves in the configuration
space of the goal object. Formally, assume we have a distance metric,
$\dist: \Cspace^g \times \Cspace^g \rightarrow \mathbb{R}^{\geq 0}$,
and a function $\fk: \Crobot \rightarrow \Cspace^g$ that computes
forward kinematics to the goal object's
configuration space.

We compute the cost of a single primitive, $a$, applied to a state $x
\in \X$ in two steps. First we compute the set $Q = \{q_1 \dots
q_{p+1}\}$ of robot configurations achieved by the primitive. This can
be obtained by forward propagating the controls in the primitive through the
constraint $f$ (\eref{eq:nonholonomic_constraint}). Then the cost of a primitive is:
\begin{equation}
\cost_a(a, x) = \sum_{i=1}^p \dist(\fk(q_i), \fk(q_{i+1}))
\end{equation}
And the cost of a path, $\pi$:
\begin{equation}
\cost_{\pi}(\pi, x_0) = \sum_{i=1}^t \cost_a(a_i, x_i)
\end{equation}
where $x_i$ is the state reached by sequentially applying primitives
$a_0 \dots a_{i-1}$ to $x_0$.

%% file: search_based.tex
\section{Heuristic Search-based Rearrangement Planning}
\label{sec:search_based}
\begin{algorithm*}[t!]
\begin{minipage}[t]{0.5\textwidth}
\begin{algorithmic}[1]
\Procedure{Search}{$x_0$}
    \State $\Basic \gets \tt{GetBasicPrimitives}$$(\Uspace)$
    \State $v \gets \tt{CreateVertex}$$(x_0, 0, \tt{ComputeHeuristic}$$(x_0))$
    \State $\Vertices.\tt{add}$$(v)$
    \State $\Queue \gets \{\}$
    \While{not $\tt{IsGoal}$$(v.x)$}
        \For{$b \in \Basic$}
           \State $\tt{ApplyPrimitive}$$(b, v, \Queue, \Vertices)$
        \EndFor
        \If{not $\tt{GoalObjectContact}$$(v.x)$}
           \State $p \gets \tt{GenerateContactPrimitive}$$(v.x)$
        \Else
           \State $p \gets \tt{GeneratePushPrimitive}$$(v.x)$
        \EndIf
        \State $\tt{ApplyPrimitive}$$(p, v, \Queue, \Vertices)$
        \State $v \gets \Queue.\tt{pop}()$
    \EndWhile
    \Return $\tt{ExtractPath}$$(v)$
\EndProcedure
\end{algorithmic}
\end{minipage}
\begin{minipage}[t]{0.5\textwidth}
\begin{algorithmic}[1]
\Procedure{ApplyPrimitive}{$p, v, \Queue, \Vertices$}
  \State $x_{new} \gets \tt{PhysicsPropagate}$$(v.x, p)$\;
  \State $g_{new} \gets v.g + \cost_a(p, v.x)$\;
  \If{not $\Vertices.\tt{find}$$(x_{new})$}
      \State $h \gets \tt{ComputeHeuristic}$$(x_{new})$\;
      \State $v_{new} \gets \tt{CreateVertex}$$(x_{new}, g_{new}, h)$\;
      \State $\Queue.\tt{push}$$(v_{new})$\;
      \State $\Vertices.\tt{add}$$(v_{new})$ \;
  \Else
      \State $v_{new} \gets \Vertices.\tt{get}$$(x_{new})$\;
      \If{$g_{new} < v_{new}.g$}
          \State $\Queue.\tt{update}$$(v_{new}, g_{new})$
      \EndIf
  \EndIf
  \Return $v_{new}$
\EndProcedure
\end{algorithmic}
\end{minipage}
\caption{A heuristic search-based planner for solving rearrangement planning.}
\label{alg:search_based}
\end{algorithm*}

Our goal is to perform an organized search across the high-dimensional
state space. We make the following three
assumptions on the planning instance:
\begin{itemize}
\item[]\textbf{\assumptionee}: Contact between the robot and goal
  object, $\goalobj$, is restricted to the end-effector. We do allow
  contact between the full robot and all other objects in $\Movables$.
\item[]\textbf{\assumptionchain}: Contact between objects is
  forbidden. For example, the robot cannot use one object to push another.
\item[]\textbf{\assumptionquasi}: All motions of, and interactions
  between, the robot and objects are quasistatic.
\end{itemize}

The search proceeds as follows. A list of vertices, $\Vertices$, each
representing a state $x \in \Xfree$ is maintained throughout the
search. The list is initialized with the start state $x_0$. At each
iteration of the search, a state is removed from the list and
expanded. During expansion, a discrete set of primitives is
applied. Each primitive is forward-propagated through a transition
model that closely approximates the non-holonomic constraint $f$.  The
resulting states are added to $\Vertices$.  The search ends when the
state removed from $\Vertices$ is a goal state.

The order that vertices are removed from $\Vertices$ is determined by the
particular search algorithm being used. Simple algorithms such as
depth-first search or breadth-first search select states based on
order of discovery. These search methods are unfocused and can lead to
unnecessary expansion of several states unlikely to be on a path to the goal.

We would like to harness the power of heuristic-based graph search
methods, such as A*. These select states based on the cost of the path
to arrive at the state and the estimated remaining cost to achieve the
goal.  In addition, the algorithm maintains a list of expanded states
and avoids unnecessary re-expansion when the same state is encountered
along a different path.

In the following sections we define the four elements needed to use
these heuristic search methods: (1) a transition model, (2) a discrete
set of primitives, (3) a heuristic function and
(4) a method for quickly determining whether a state has been expanded
previously.

With these four elements defined we can use any heuristic
search-based method to perform the search~\cite{hart68astar,
  likhachev04ara,pohl70weightedastar,aine14mhastar}.~\aref{alg:search_based}
outlines how our primitives, heuristics and mappings can be applied to
an A*-based algorithm.

\begin{figure}
\centering
\includegraphics[width=0.55\linewidth]{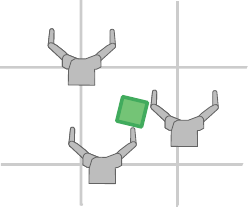}
\caption{A simple example of a hand pushing objects.  The
  \textit{basic} primitives allow the hand to translate along the grid
  lines. In this simple example, the object is ``trapped'', i.e. there
  are no primitives that allow it to move out of the cell.}
\label{fig:toy_example}
\end{figure}

\subsection{Transition Model}
\label{sec:transition_model}
We require a transition model that closely approximates $f$
(\eref{eq:nonholonomic_constraint}). We use a quasistatic pushing
model with Coulomb friction~\cite{lynch95stablepushing}. In this
model we assume pushing motions are slow enough that inertial forces
are negligible; objects only move when pushed by the robot and
objects stop immediately when forces cease to be imparted on the
object.

\subsection{Action Selection}
We select a discrete set of actions, or primitives, to apply to each
state that allow us to perform a feasible and focused search. We first
select a set of primitives that move the robot without the explicit
intent of creating contact or interaction with objects. These
primitives are small motions of the robot defined by a coarse
discretization of the control space. We label these primitives
\textit{basic} primitives.

These \textit{basic} primitives are context agnostic; they are
not specific to the rearrangement planning problem. We know contact
with the goal object is critical to goal achievement. 
\textit{Basic} primitives may achieve some contact with objects, but
it is not guaranteed.  Consider the simple example of a robot pushing
an object in~\figref{fig:toy_example}. The \textit{basic} primitives
move in the four cardinal directions. The object is ``trapped'' ,
i.e. there is no primitive that can create enough contact to move it
out of the current cell.

We could expand the set of \textit{basic} primitives by discretizing
the control space more finely. However the size
of the primitive set is directly related to the branching factor of
the search, so any large increase affects the computation time.

Instead we propose to augment the primitive set applied at each state
with a dynamically-generated action aimed at creating or maintaining
contact with the goal object. We label these primitives as
\textit{contact} and \textit{pushing} primitives, respectively. The
use of these primitives focuses our search to areas of the full state
space that are likely to lead to the goal.

We observe that it is often possible to solve the 2PTBVP in the
lower-dimensional subspace containing only the robot. We use this to
design the \textit{contact} and \textit{pushing} primitives.

To generate a \textit{contact} primitive to be applied to a state $x
\in \Xfree$, we first find a pose, $q_{con}$, of the robot such that the
end-effector is in or near contact with the goal object in
configuration $o^g \in x$. Then we solve the 2PTBVP in the robot
configuration space to generate motions that move from $q$ to
$q_{con}$. During the search, we apply a \textit{contact} primitive to
any state where the robot and goal object are not in or near contact.

A \textit{pushing} primitive aims to push the goal object toward the
goal region. During the search we apply a \textit{pushing} primitive
to any state where the robot and goal are in or near contact. To generate the
primitive we produce any motion of the robot that moves in the
direction of the goal object. We detail a specific example \textit{contact}
and \textit{pushing} primitive in our experiments
in~\sref{sec:results}.

The \textit{basic} and \textit{contact} primitives are similar to
transit actions defined by Simeon et al.~\cite{simeon04prm}, where the
robot changes configuration without moving an object.  The
\textit{pushing} primitive mimics transfer actions, where an object is
grasped and reconfigured.  We note that the correspondence is not
exact, \textit{basic} and \textit{contact} primitives may
inadvertently make contact with objects in the scene. Likewise,
\textit{pushing} primitives may lose contact with the goal object in
the middle of primitive execution.

\subsection{Heuristic}


Developing a heuristic that estimates the distance between a state and
the goal region is challenging because the goal is
underspecified. In particular, the goal is defined only by the configuration of
$\goalobj$. The configuration of the robot is not defined.  As a
result, common robot-configuration based heuristics are not directly
applicable.

However, two observations of the problem can be used to generate a
useful heuristic. First, by definition of the problem, contact with
the goal object is required for goal achievement. Due to Assumptions 1
and 2, this contact must be between the end-effector and the
object. Second, the robot must stay in contact with the goal object
for the object to move, due to \assumptionquasi.

Using these observations we develop a two part heuristic to estimate
the distance between a state $x \in \Xfree$ and $\Xgoal$:
\begin{align}
h(x) =&~\chatcontact(x) + \label{eq:heuristic1} \\ 
&~\chatmove(x) \label{eq:heuristic2} 
\end{align}
\eref{eq:heuristic1} estimates the distance to make contact with the
goal object.~\eref{eq:heuristic2} estimates the distance the
end-effector must move to push the goal object to the goal region.
\newline\newline
\noindent\textit{\textbf{Distance to contact:}} We compute $\chatcontact$ by
approximating the end-effector with the smallest enclosing sphere. If
this sphere penetrates the object, $\chatcontact=0$. Otherwise,
$\chatcontact$ is the translational distance between the closest points on the
sphere and object under the metric $\dist$.

\begin{proposition}
$\chatcontact$ is a lower bound on the true cost to make contact with
  the goal object.
\end{proposition}
\begin{proof}
  Approximating the end-effector pose with the sphere
  means all rotations of the end-effector have
  $\chatcontact=0$. Thus our estimate of the rotation distance is
  a lower bound of the true distance. The shortest translational distance
  the end-effector can move to make contact is the distance between
  the two closest points on the end-effector and object. Selecting the
  closest point on the sphere to the object ensures we underestimate
  this distance. Since we underestimate translational and rotational
  distance, we must underestimate the true distance.
\end{proof}
~\newline
\noindent\textit{\textbf{Distance to goal: }} We compute $\chatmove = \dist(o^g,
p_{goal}) - r_{goal}$. This is the straight line distance from the
object location to the edge of the goal region. 
\begin{proposition}
$\chatmove$ is a lower bound on the true cost to
  move the object to the goal.
\end{proposition}
\begin{proof}
$\chatmove$ is the shortest distance the
  object can move and still achieve the goal. By the quasistatic
  assumption, contact must be maintained between robot and object for
  the object to move. As a result, $\chatmove$ must also be the
  shortest distance the robot could move. 
\end{proof}

\subsection{State equality}
\begin{figure}[t!]
\captionsetup[subfigure]{margin=5pt}
\begin{subfigure}[t]{0.5\linewidth}
\centering
\includegraphics[width=\linewidth]{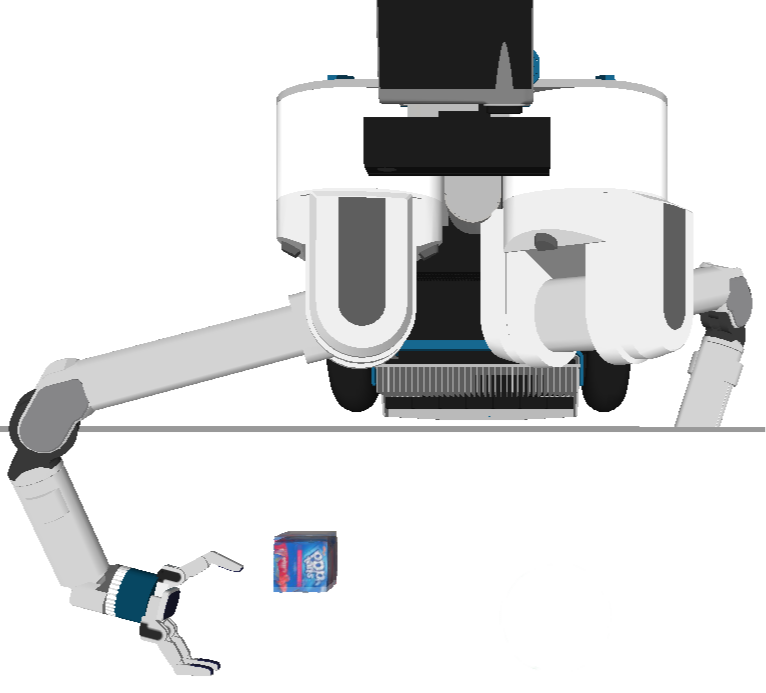}
\caption{Two states with object pose that differs by less than 1cm}
\label{fig:start_combined}
\end{subfigure}%
\begin{subfigure}[t]{0.5\linewidth}
\centering
\includegraphics[width=\linewidth]{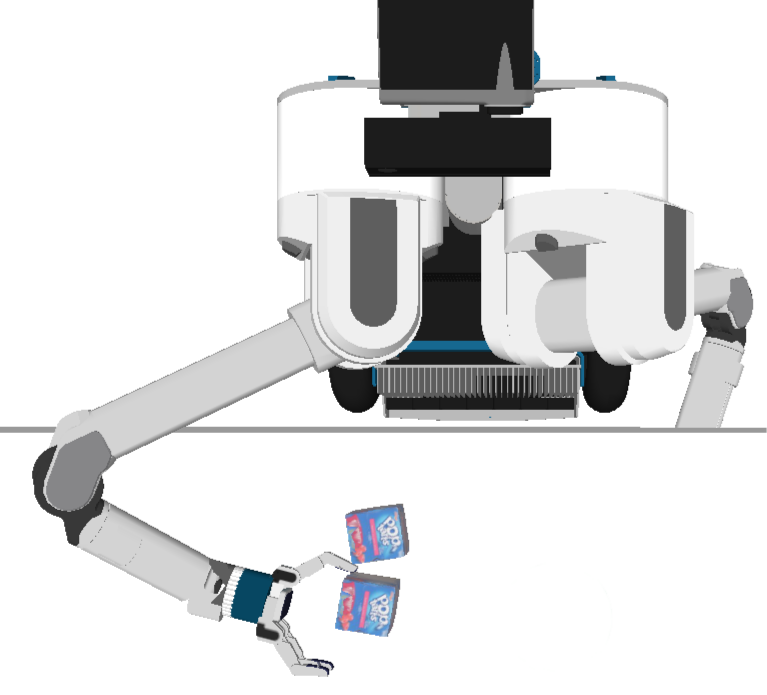}
\caption{After executing the same action, the final object pose differs significantly.}
\label{fig:end_combined}
\end{subfigure}
\caption{Under typical mappings from continuous to discrete states, the two states
  depicted in (\protect\subref{fig:start_combined}) would be labeled the
  same. However, applying the same primitive leads to very different
  final states (\protect\subref{fig:end_combined}).}
\label{fig:state_mapping}
\end{figure}
To more effectively use heuristic-based search methods we need to
recognize when the search has encountered the same state along two
different paths. One commonly used method is to define a mapping from
a state $x \in \Xfree$ to a discrete bin. Any two states that map to
the same discrete bin are considered equivalent.

This method relies on the assumption that two states that start near
each other will end near each other when a primitive is applied. The
discontinuous nature of pushing interactions means this assumption
often does not hold. Consider the example
from~\figref{fig:state_mapping}. A very small change in initial
object pose~(\figref{fig:start_combined}) leads to large difference of
final pose~(\figref{fig:end_combined}), despite the same primitive
being applied. As a result we can only mark two states as
equivalent when they are exactly equivalent. To do this, we track all
states encountered along the search and use a Geometric Near-neighbor
Access Tree (GNAT)~\cite{brin95gnat} to efficiently find the nearest
neighbor to a new state and check equality.

%% file: results.tex
\section{Experiments and Results}
\label{sec:results}

We have implemented our planner using the Boost Graph Library~\cite{bgl}.  Using
the planner we test three hypotheses:
\begin{enumerate}
\item[] \hypothesis{H1}: The use of \textit{contact} and
  \textit{pushing} primitives improves success rate when compared to a
  planner that uses only \textit{basic} primitives.
\item[] \hypothesis{H2}: The use of the search-based planner is able to
  produce more optimal paths than current randomized methods.
\item[] \hypothesis{H3}: The goal-directed nature of the
  \textit{contact} and \textit{pushing} primitives allows the search
  planner to find solutions in plan times comparable to current randomized methods.
\end{enumerate}
The experiments are conducted using a 7 degree-of-freedom arm, first
in simulation and then on the real robot.

\subsection{Experiment Setup}
We evaluate our hypotheses on 12 scenes, each containing between 1
and 7 movable objects.~\figref{fig:simulation_filmstrip} shows an
example scene. We use the same robot start configuration and goal
region in all 12 scenes. Each scene contains the same goal object,
but its start location, and the number and start location of all other
objects differs across scenes. The objects are placed on a table. The
table is treated as an obstacle that the robot is forbidden to
contact. We set $r_{goal} = 10$cm for all scenes.  We use a
quasistatic model of physics to propagate interactions between the
objects and the robot. We model only planar pushing actions (no
toppling) for the 7-DOF arm. Thus $\Crobot = \mathbb{R}^7$ and
$\Cspace^1 = \dots = \Cspace^m = \setwo$. We follow the ideas
from~\cite{king15nonprehensile} and constrain the motion of the
end-effector to the plane parallel to the pushing support surface. As
required by \assumptionee~we invalidate any robot motions that create
contact between the goal object and any part of the robot other than
the manipulator. We note that we allow and model pushing contact
between the full arm and all objects except the goal object. As
required by \assumptionchain, we also invalidate any robot motions
that lead objects to contact one another.


The planner is given a set of 6 \textit{basic} primitives.  Each of
these six actions moves the end-effector in $SE(2)$ at a pre-defined
maximum velocity (positive or negative) along a single axis ($x$, $y$,
$\theta$).  We use a Jacobian psuedo-inverse controller to compute
the 7-DOF velocities that achieve the desired end-effector motion. The
maximum linear (\SI{0.5}{m/s}) and angular (\SI{1.0}{rad/s}) velocity limits were
selected to ensure the resulting 7-DOF joint velocities remained
within safety limits defined for the robot.  We set the duration of
each \textit{basic} primitive to \SI{0.2}{s}.

\begin{figure}
\captionsetup[subfigure]{margin=5pt}
\begin{subfigure}[b]{0.5\linewidth}
\centering
\includegraphics[width=0.98\linewidth]{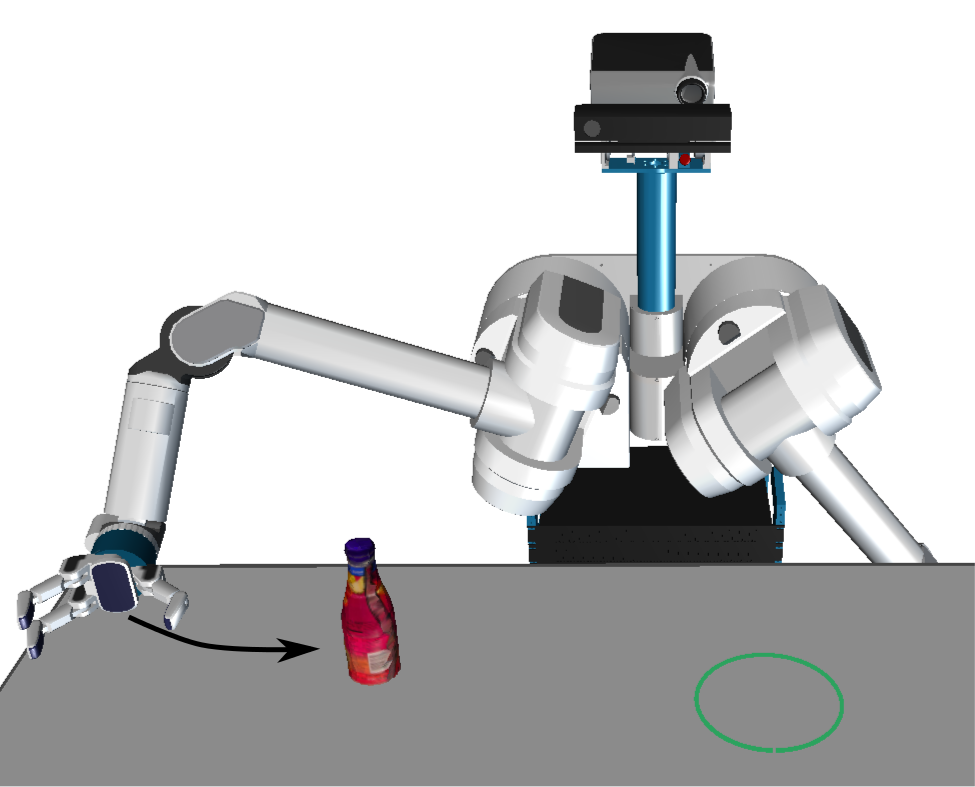}
\caption{The \textit{contact} primitive tries to make contact between
  the palm and the goal object.}
\label{fig:contact_primitive}
\end{subfigure}%
\begin{subfigure}[b]{0.5\linewidth}
\centering
\includegraphics[width=0.98\linewidth]{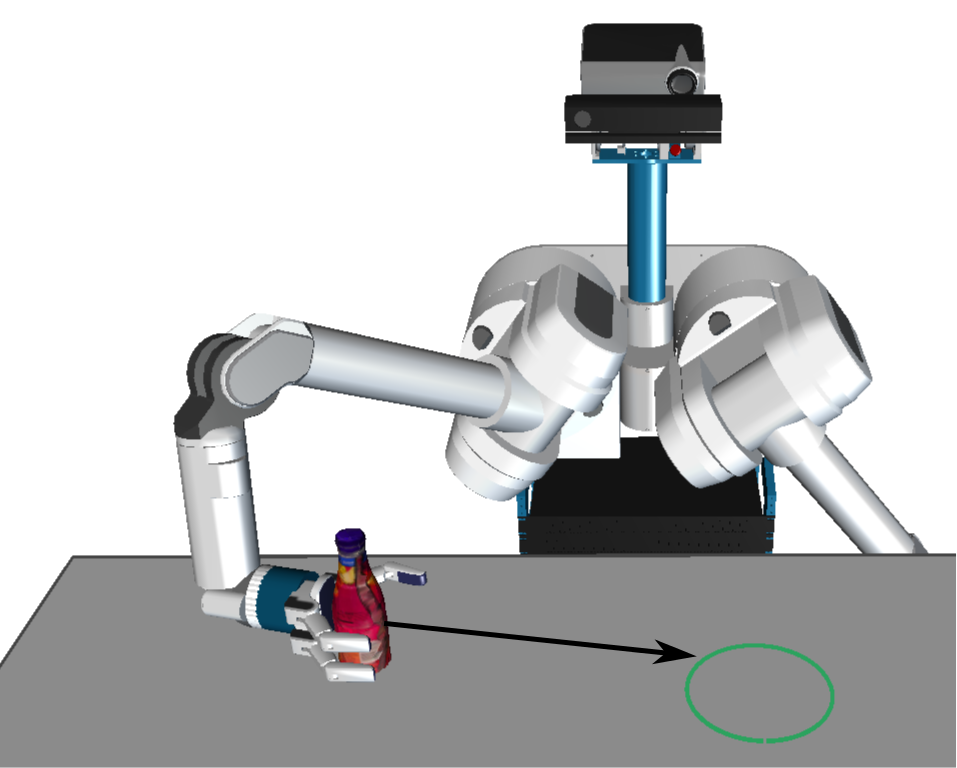}
\caption{The \textit{pushing} primitive moves the end-effector in the
  direction of the goal region}
\label{fig:pushing_primitive}
\end{subfigure}%
\caption{Context specific primitives are used to guide the search.   }
\label{fig:manipulator_primitives}
\end{figure}

The \textit{contact} primitive defined on an object moves the
end-effector along a straight line in $SE(2)$ toward the goal object
until either contact is made or an invalid state is
encountered. During the motion, the end-effector is rotated to face
the palm of the hand to the object centroid. The \textit{pushing} primitive
moves the end-effector in a straight line in $SE(2)$ along the vector
normal to the palm.~\figref{fig:manipulator_primitives} depicts an
example \textit{contact} and \textit{pushing} primitive.

\subsubsection{Baseline Planners}
We compare the performance of our search-based planner (denoted
\pushingsearch in all results) against three baseline
planners.  First we compare against a planner that uses only the
\textit{basic} primitives defined above. We denote this planner as
\basicsearch in all results and discussions.

Next, we compare against the RRT planner used
in~\cite{king15nonprehensile}.  This planner grows a search tree from
the start state toward a goal region by iteratively sampling a
random state from $\Xfree$, and growing the tree toward the
sample. During tree extension we sample $k=3$ actions and use the
quasistatic model of physics to forward-simulate each action. We
keep the action that ends closest to the sample under a
weighted euclidean distance metric.  We denote this planner as
\pushingrrt in all results and discussions.

Finally, we compare against an altered version of the RRT that
includes manipulation primitives similar to our \textit{contact} and
\textit{pushing} primitives. In this planner, during tree extension, a
\textit{contact} primitive is applied to move the robot near an
object, then a \textit{pushing} primitive is applied to move the
object. The moved object and push direction are indicated in the
sampled state the tree is growing towards. We denote this planner as
\primitiverrt in all results and discussions.

For the RRT-based planners, we do not apply \assumptionee~and allow
contact between the full robot and the goal object. Our quasistatic
physics model does not model object-to-object interaction. As a result we
must still impose \assumptionchain~on the RRT planners.

\subsection{Analysis}

\begin{figure*}
\captionsetup[subfigure]{margin=5pt}
\begin{subfigure}[t]{0.24\textwidth}
\centering
\includegraphics[width=\textwidth]{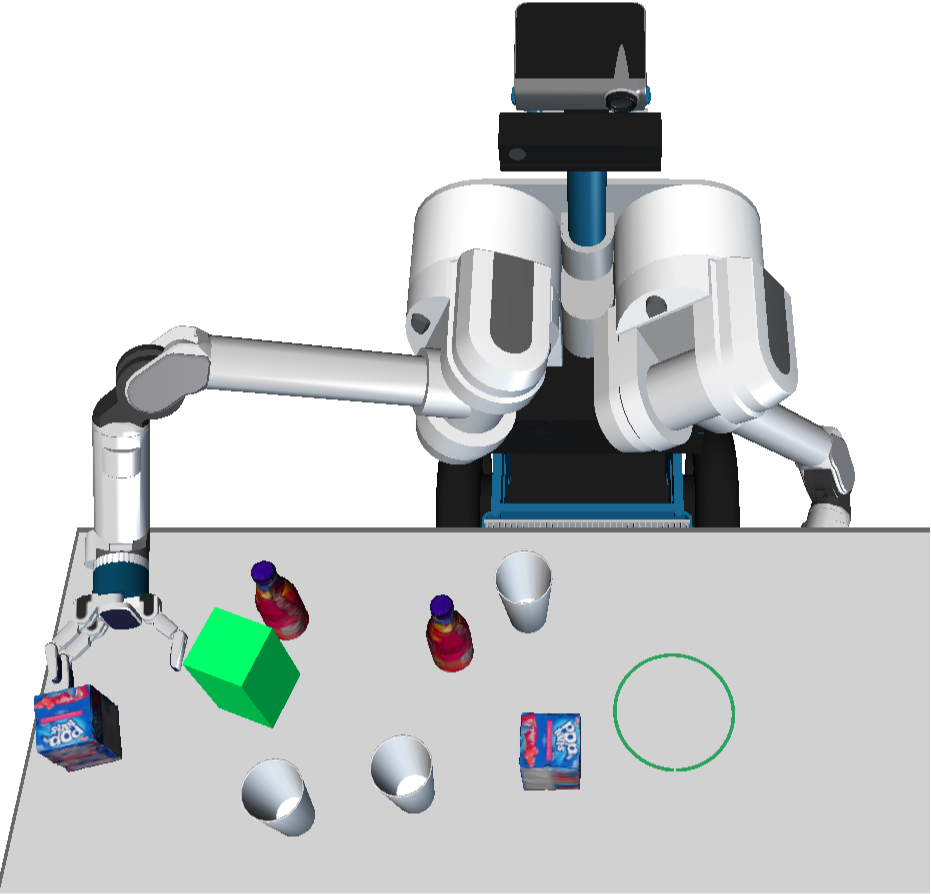}
\caption{In many paths generated by the \pushingsearch, the robot's
  initial motion is a \textit{contact} primitive.}
\label{fig:filmstrip_1}
\end{subfigure}
\begin{subfigure}[t]{0.24\textwidth}
\centering
\includegraphics[width=\textwidth]{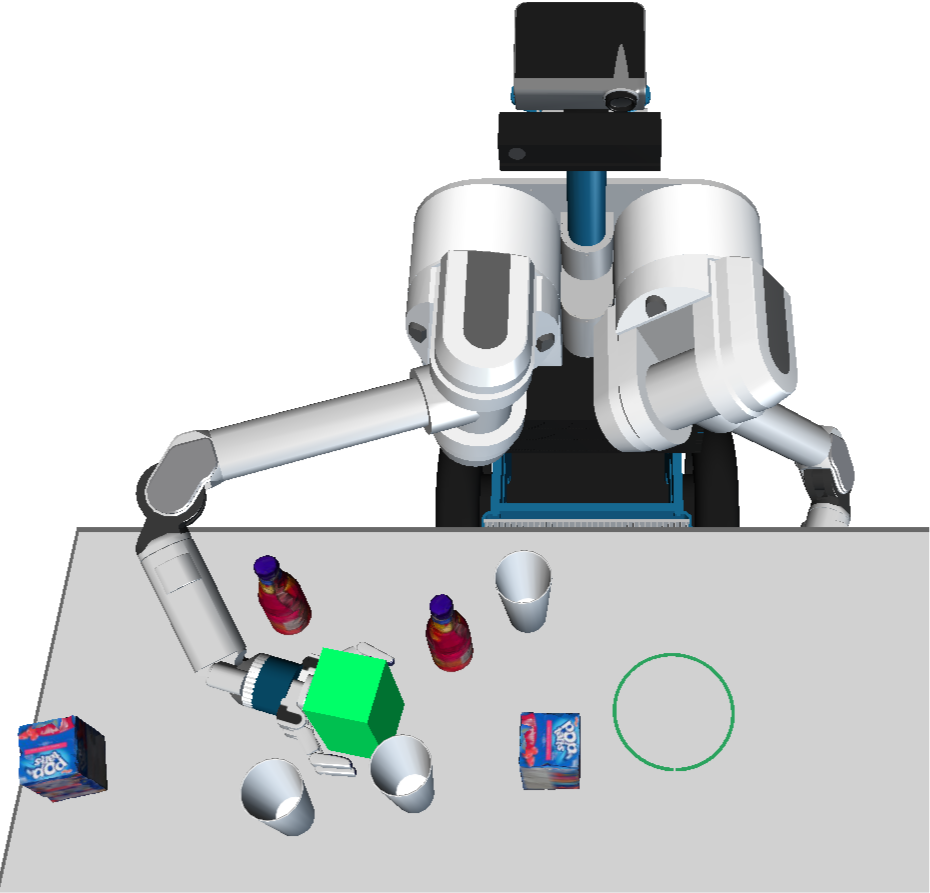}
\caption{\textit{Basic} primitives are used to reposition the robot
  when stuck. Here the box cannot be pushed further without contacting
  the glass.}
\label{fig:filmstrip_regrasp1}
\end{subfigure}
\begin{subfigure}[t]{0.24\textwidth}
\centering
\includegraphics[width=\textwidth]{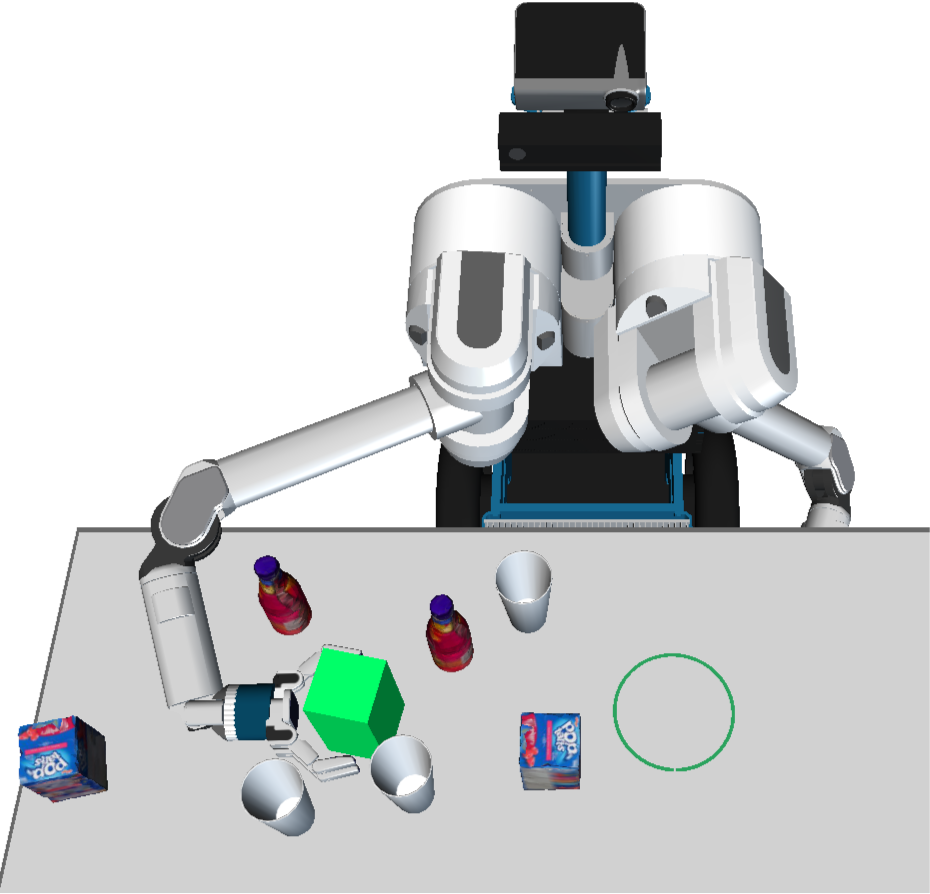}
\caption{After the \textit{basic} rotation primitive reorients the
  arm, the robot can continue a \textit{pushing} primitive towards the
  goal.}
\label{fig:filmstrip_regrasp2}
\end{subfigure}
\begin{subfigure}[t]{0.24\textwidth}
\centering
\includegraphics[width=\textwidth]{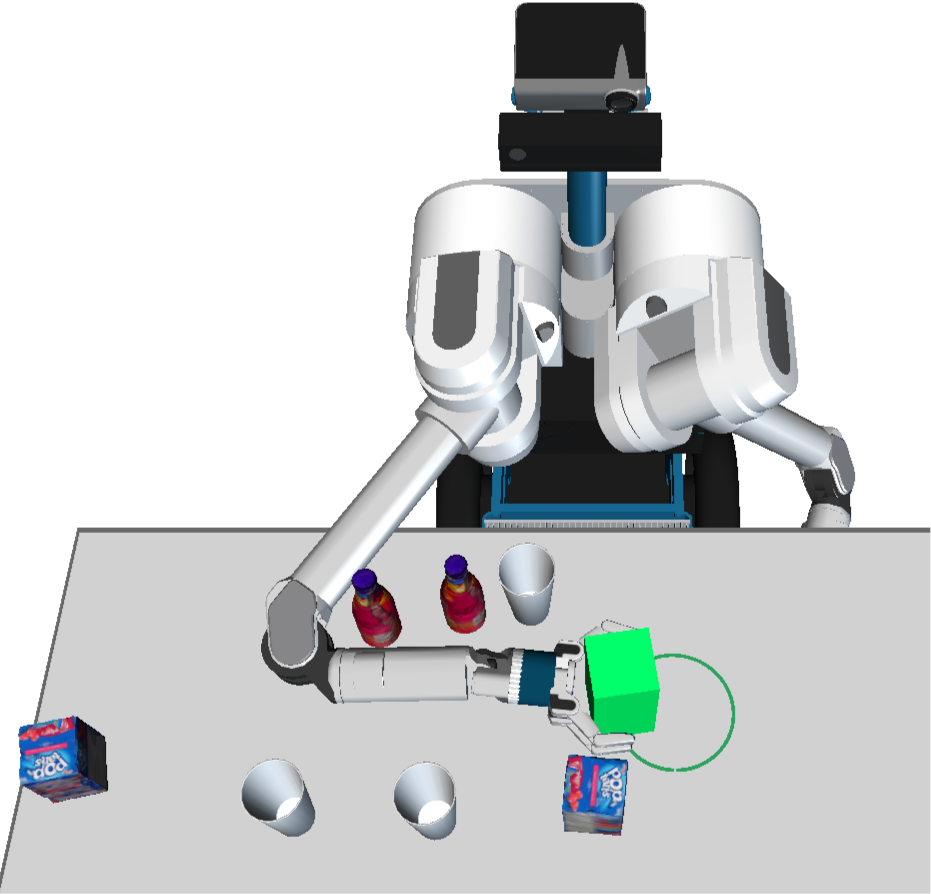}
\caption{As the object is pushed to the goal, the fingers and upper
  arm are used to push clutter aside.}
\label{fig:filmstrip_goal}
\end{subfigure}
\caption{A solution for one of the 12 simulation scenes. The robot is
  tasked with pushing the green box into the green circle.}
\label{fig:simulation_filmstrip}
\end{figure*}

\subsubsection{Effect of Manipulation Primitives}
We first compare the performance of the \pushingsearch planner against
the \basicsearch planner.  We run each planner on the 12
scenes. We allow the planners 300s to find a solution for each
scene.  If a solution is not returned within this time the run is
marked failure. We use a Weighted A*~\cite{pohl70weightedastar} search
with $w=5.0$ in all searches.

\begin{figure}
\includegraphics{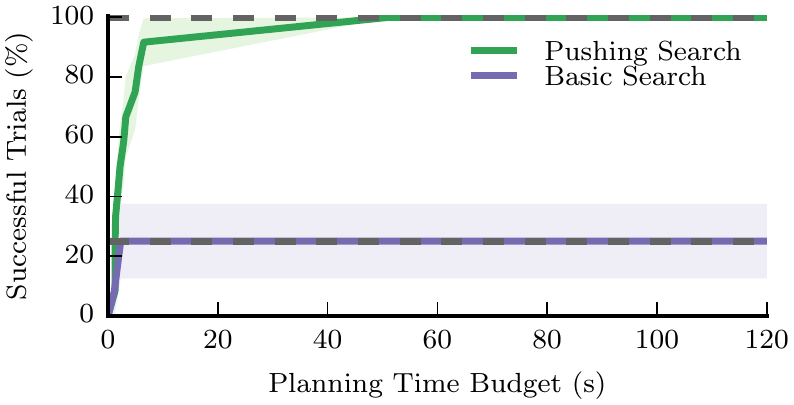}
\caption{Percent of successfully solved scenes as a function of
  plan time. Shaded regions represent standard error.}
\label{fig:herb_basic_manip_time_success}
\end{figure}

~\figref{fig:herb_basic_manip_time_success} shows the percent of scenes
solved successfully as a function of plan time. When
using only \textit{basic} primitives, the search-based planner is only
able to solve 3 of the 12 scenes within the time limit.  The
addition of the \textit{contact} and \textit{pushing} primitives
allows the planner to solve all 12 scenes in the 300s time limit.
This supports \hypothesis{H1}: \textbf{The use of
  \textit{contact} and \textit{pushing} actions improves success rate
  of search-based planners.}

\subsubsection{Path Optimality}
Next we compare the optimality of the paths produced by the
\pushingsearch planner against the paths produced by the
\pushingrrt and the \primitiverrt.  We run the
randomized planners 30 times on each scene, allowing each planner up
to 300s to find a solution. We compute the path length as the average
across all successfully planned paths.

\begin{figure}[t]
\captionsetup[subfigure]{margin=5pt}
\savebox{\largestimage}{\includegraphics{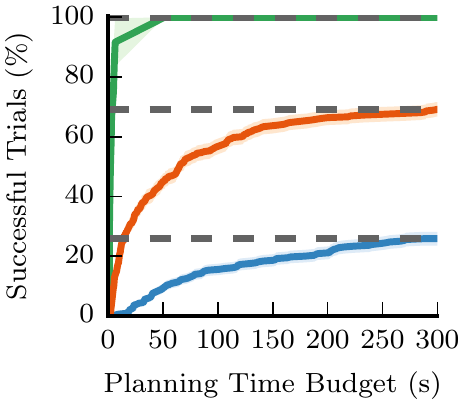}}
\begin{subfigure}[t]{.5\linewidth}
  \centering
  \raisebox{\dimexpr.55\ht\largestimage-.5\height}{\includegraphics{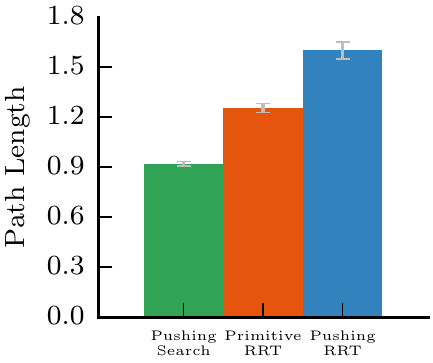}}
  \caption{A comparison of average path length across all scenes.}
\label{fig:herb_path_lengths}
\end{subfigure}%
\begin{subfigure}[t]{.5\linewidth}
  \centering
  \usebox{\largestimage}
  \caption{The percent of successful trials as a function of plan
    time. Shaded regions represent standard error.}
  \label{fig:herb_search_rrt_plan_time}
\end{subfigure}
\caption{Performance of the \pushingrrt as compared to RRT based
  planners that solve rearrangement problems}
\label{fig:herb_search_rrt}
\end{figure}

~\figref{fig:herb_path_lengths} shows a comparison of the average path
length for the three planners. A one-way ANOVA with Tukey HSD post-hoc
analysis reveals that the path length for paths produced by our
\pushingsearch algorithm is significantly smaller that the average path
lengths for paths produced by the \pushingrrt ($p=0.00$) and the
\primitiverrt ($p=0.00$). This supports our second
hypothesis: \hypothesis{H2}: \textbf{Our search-based planner with
  \textit{contact} and \textit{pushing} primitives is able to produce
  more optimal paths than randomized planners.}

%

\subsubsection{Plan Time}
~\figref{fig:herb_search_rrt_plan_time} compares the percent of successful
trials as a function of plan time for the \pushingsearch planner to
that of the \pushingrrt and \primitiverrt. As can be seen, the
\pushingsearch is able to solve more trials faster than either RRT
method. This supports our third hypothesis: \hypothesis{H3}: \textbf{
  The use of goal-focused primitives allows the search based planner
  to find solutions faster than the randomized planners.}

These results may appear surprising because randomized planners, and in
particular bi-directional randomized planners, have been shown to be
faster than search-based methods in high-dimensional spaces. Because
we cannot solve the 2PTBVP, both the \pushingrrt and \primitiverrt
must be implemented as single-directional planners where only one tree
is grown rooted at the start state.  Additionally, extensions to the
trees fail to extend all the way to the sampled state, especially in
the \pushingrrt. As a result, the planners lack the ability to perform
focused extension toward the goal. Conversely, the \textit{contact} and
\textit{pushing} are explicitly designed to lead the search toward
the goal. We believe this accounts for the fast planning times exhibited
by the planner on our test scenes.

\subsection{Real Robot Experiments}
\begin{figure*}[ht!]
\captionsetup[subfigure]{margin=5pt}
\begin{subfigure}[b]{0.2\linewidth}
\centering
\includegraphics[width=0.98\linewidth]{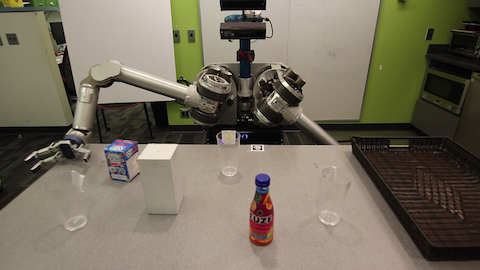}
\end{subfigure}%
\begin{subfigure}[b]{0.2\linewidth}
\centering
\includegraphics[width=0.98\linewidth]{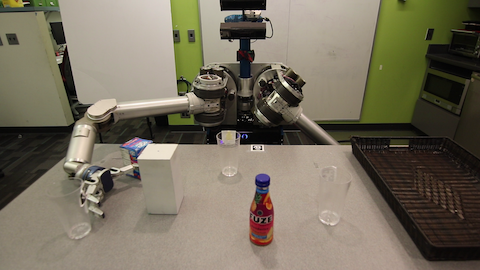}
\end{subfigure}%
\begin{subfigure}[b]{0.2\linewidth}
\centering
\includegraphics[width=0.98\linewidth]{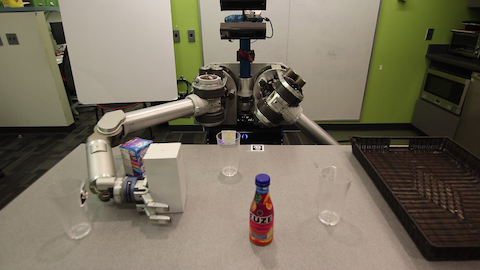}
\end{subfigure}%
\begin{subfigure}[b]{0.2\linewidth}
\centering
\includegraphics[width=0.98\linewidth]{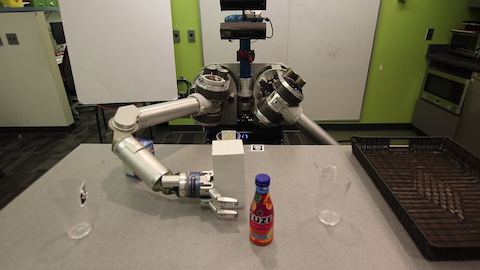}
\end{subfigure}%
\begin{subfigure}[b]{0.2\linewidth}
\centering
\includegraphics[width=0.98\linewidth]{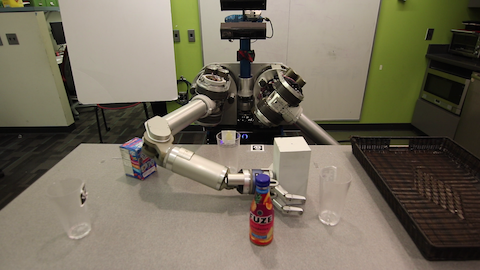}
\end{subfigure}
\caption{The \pushingsearch planner is able to find a path using a single \textit{contact} primitive followed by a single \textit{pushing} primitive.}
\label{fig:filmstrip_scene3}
\end{figure*}

Next we perform a set of trials on the real robot. We use the planners
to generate plans for 12 scenes similar to those used in
simulation. For each scene we randomly place between 1 and 6 objects
on a table within the reachable workspace of the robot. The pose of
these objects is detected using AprilTag fiducial
markers~\cite{olson11tags}.~\figref{fig:figure1},~\figref{fig:filmstrip_scene10}
and~\figref{fig:filmstrip_scene3} depict 3 of the 12 scenes.

We use the same parameters for the
\pushingsearch algorithm as used in the simulation results. All 12
scenes have the same start configuration for the robot and goal
region. We set $r_{goal} = 10$cm across
all scenes.

\begin{figure}
\captionsetup[subfigure]{margin=5pt}
\savebox{\largestimage}{\includegraphics{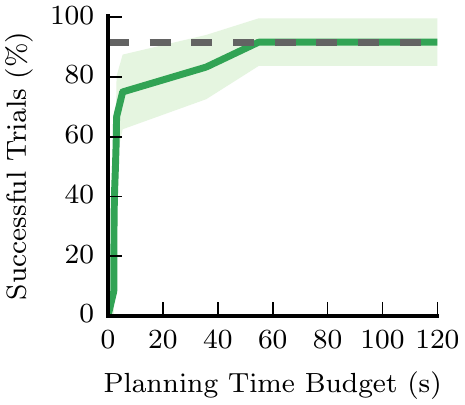}}

\begin{subfigure}[t]{0.5\linewidth}
  \centering
  \raisebox{\dimexpr.7\ht\largestimage-.5\height}{\includegraphics{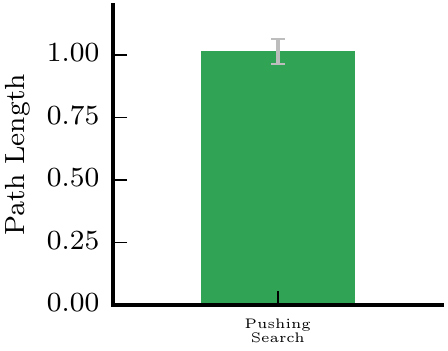}}
  \caption{The lengths of the paths generated by the
    \pushingsearch planner.}
  \label{fig:real_path_length}
\end{subfigure}%
\begin{subfigure}[t]{0.5\linewidth}
  \centering
  \usebox{\largestimage}
  \caption{The percent scenes successfully solved by the
    \pushingsearch planner as a function of plan time. Shaded regions
    represent standard error.}
  \label{fig:real_robot_time_success}
\end{subfigure}
\caption{Results from trials on 12 scenes for the real robot.}
\end{figure}

For each trial we record the time to generate a plan using our
planner.~\figref{fig:real_robot_time_success} shows the percent of
scenes solved as a function of plan time. The planner is able to
successfully find solutions for 11 of the 12 scenes.

\subsection{Qualitative Analysis}
Comparing the paths across the 24 scenes (12 simulation, 12
real robot) reveals some interesting qualitative aspects of the
solutions. First, the \textit{contact} and \textit{pushing} primitives
comprise the majority of the paths.  This is not surprising as they
have the most effect on the heuristic cost. In particular, a
successfully applied \textit{contact} primitive
reduces~\eref{eq:heuristic1} to zero while the \textit{pushing}
primitive reduces~\eref{eq:heuristic2}. The result is that
the planner is guided to create paths that move directly to the goal
object then push it almost directly to the
goal~(\figref{fig:simulation_filmstrip},~\figref{fig:filmstrip_scene3}),
using the fingertips, back of the hand and arm to clear clutter along
the way.

In these solutions, we see \textit{basic} primitives used as
``regrasp'' motions. Consider the motion
between~\figref{fig:filmstrip_regrasp1}
and~\figref{fig:filmstrip_regrasp2}. In~\figref{fig:filmstrip_regrasp1}
the green box cannot be pushed further without hitting the glass, an
action that violates \assumptionchain~. A \textit{basic} rotation
primitive is used to reorient the robot, allowing further pushing of
the object.

%% file: discussion.tex
\section{Discussion}
\label{sec:discussion}
In this work, we formulate a method for applying heuristic search to
rearrangement planning by pushing. We show the use of dynamically
generated, problem specific primitives that are critical to goal
achievement. These primitives, combined with an informative and
admissible heuristic, guide the search to promising areas of state
space. Our experiments show we are able to quickly produce low cost
paths for several problems.

While our experiments are promising, our formulation imposes several
limitations to address in future work. We define the cost function we
optimize in the configuration space of the goal object. This
simplifies the definition and computation of $\chatcontact$
(\eref{eq:heuristic1}) and $\chatmove$
(\eref{eq:heuristic2}). However, it is often more desirable to express
cost in the configuration space of the robot, e.g. path length in
joint space.  For manipulators such as the one used for our
experiments, it is difficult to compute meaningful and admissible
estimates of the distance in joint space to contact the goal
object and the minimum joint motions that move the object. However, if we
could find a computationally tractable method for estimating these
values, we could not only use a more desirable cost function, but we
could also eliminate \assumptionee~. This would allow solutions that
exhibit whole-arm or even whole-body interactions with the goal
object.

\assumptionchain~prevents solutions where the robot uses one object to
push another. This is particularly restrictive when the robot is
working in highly cluttered spaces. We could eliminate this assumption
at the expensive of a less informed heuristic.  In particular, we
could define $\chatcontact$ as the distance to the \textit{nearest}
object rather than distance to the goal object.  This implies that the
robot can push the nearest object and create a chain of pushes that moves
the goal object. This will often grossly underestimate the true
distance to goal. Future work should examine whether a tighter bound
could be inferred from the structure of the state.

While the quasistatic assumption is common in planning pushing
interactions~\cite{mason86plan, peshkin88slide,
  lynch95controllability, lynch95stablepushing, lynch97locallycontrol,
  akella92polygonal, dogar10pushgrasp, nieuwenhuisen05discpush},
incorporating dynamic interactions increases the space of possible
solutions. For example, the robot could simply move to the object,
strike it and remain in place as the object slides to the
goal~\cite{haustein15kinodynamic,zickler09physics}. A naive
incorporation of dynamic interactions into the heuristic sets
$\chatmove = 0$. This is a looser bound than we are able to formulate
under the quasistatic assumption. Additionally, we incur the penalty
of a state space that doubles in size, as we must track the
configurations \textit{and} velocities of all objects. 

Despite these limitations we believe the results presented here are
promising. The methods for selecting a discrete set of primitives from
the continuous space of robot motions can be applied to online
planning and robust path generation for the rearrangement
problem. This will allow planners to represent and reason about
uncertainty and incorporate feedback, resulting in better execution of
rearrangement plans.

%% file: acknowledgments.tex
\section*{Acknowledgments}

%% file: RSS2016-rearrangement.bbl
\begin{thebibliography}{36}
\providecommand{\natexlab}[1]{#1}
\providecommand{\url}[1]{\texttt{#1}}
\expandafter\ifx\csname urlstyle\endcsname\relax
  \providecommand{\doi}[1]{doi: #1}\else
  \providecommand{\doi}{doi: \begingroup \urlstyle{rm}\Url}\fi

\bibitem[bgl(2002)]{bgl}
\emph{The Boost Graph Library: User Guide and Reference Manual}.
\newblock Addison-Wesley Longman Publishing Co., Inc., 2002.
\newblock ISBN 0-201-72914-8.

\bibitem[Aine et~al.(2014)Aine, Swaminathan, Narayanan, Hwang, and
  Likhachev]{aine14mhastar}
S.~Aine, S.~Swaminathan, V.~Narayanan, V.~Hwang, and M.~Likhachev.
\newblock Multi-heuristic {A}*.
\newblock In \emph{{RSS}}, 2014.

\bibitem[Akella and Mason(1992)]{akella92polygonal}
S.~Akella and M.~T. Mason.
\newblock Posing polygonal objects in the plane by pushing.
\newblock In \emph{{IEEE} {ICRA}}, 1992.

\bibitem[Barry et~al.(2012)Barry, Hsiao, Kaelbling, and
  Lozano-P\'{e}rez]{barry12darrt}
J.~Barry, K.~Hsiao, L.~P. Kaelbling, and T.~Lozano-P\'{e}rez.
\newblock Manipulation with multiple action types.
\newblock In \emph{{ISER}}, 2012.

\bibitem[Brin(1995)]{brin95gnat}
S.~Brin.
\newblock Near neighbor search in large metric spaces.
\newblock In \emph{Proceedings of the 21st International Conference on Very
  Large Data Bases}, 1995.

\bibitem[Chen and Hwang(1992)]{chen98sandros}
P.C. Chen and Y.K Hwang.
\newblock {SANDROS}: A dynamic graph search algorithm for motion planning.
\newblock In \emph{{IEEE} {ICRA}}, 1992.

\bibitem[Cohen et~al.(2010)Cohen, Chitta, and
  Likhachev]{cohen10motionprimitives}
B.~Cohen, S.~Chitta, and M.~Likhachev.
\newblock Search-based planning for manipulation with motion primitives.
\newblock In \emph{{IEEE} {ICRA}}, 2010.

\bibitem[Dogar and Srinivasa(2010)]{dogar10pushgrasp}
M.R. Dogar and S.S. Srinivasa.
\newblock Push-grasping with dexterous hands: Mechanics and a method.
\newblock In \emph{{IEEE/RSJ} {IROS}}, 2010.

\bibitem[Dogar and Srinivasa(2011)]{dogar11pushgrasping}
M.R. Dogar and S.S. Srinivasa.
\newblock A framework for push-grasping in clutter.
\newblock In \emph{{RSS}}, 2011.

\bibitem[Gammell et~al.(2014)Gammell, Srinivasa, and Barfoot]{gammell14irrt}
J.D. Gammell, S.~Srinivasa, and T.D. Barfoot.
\newblock Informed {RRT}*: Optimal sampling-based path planning focused via
  direct sampling of an admissible ellipsoidal heuristic.
\newblock In \emph{{IEEE/RSJ} {IROS}}, 2014.

\bibitem[Gammell et~al.(2015)Gammell, Srinivasa, and Barfoot]{gammell15bit}
J.D. Gammell, S.~Srinivasa, and T.D. Barfoot.
\newblock Batch informed trees ({BIT}*): Sampling-based optimal planning via
  the heuristically guided search of implicit random geometric graphs.
\newblock In \emph{{IEEE} {ICRA}}, 2015.

\bibitem[Hart et~al.(1968)Hart, Nillson, and Raphael]{hart68astar}
P.E. Hart, N.J. Nillson, and B.~Raphael.
\newblock A formal basis for the heuristic determination of minimum cost paths.
\newblock In \emph{{TSSC}}, 1968.

\bibitem[Hauser and Ng-Thow-Hing(2010)]{hauser10shortcut}
K.~Hauser and V.~Ng-Thow-Hing.
\newblock Fast smoothing of manipulator trajectories using optimal
  bounded-acceleration shortcuts.
\newblock In \emph{{IEEE} {ICRA}}, 2010.

\bibitem[Haustein et~al.(2015)Haustein, King, Srinivasa, and
  Asfour]{haustein15kinodynamic}
J.A. Haustein, J.E. King, S.S. Srinivasa, and T.~Asfour.
\newblock Kinodynamic randomized rearrangement planning via dynamic transitions
  between statically stable configurations.
\newblock In \emph{{IEEE} {ICRA}}, 2015.

\bibitem[Karaman and Frazzoli(2011)]{karaman11optmotplan}
S.~Karaman and E.~Frazzoli.
\newblock Sampling-based algorithms for optimal motion planning.
\newblock \emph{{IJRR}}, 30\penalty0 (7):\penalty0 846--894, 2011.

\bibitem[King et~al.(2015)King, Haustein, Srinivasa, and
  Asfour]{king15nonprehensile}
J.E. King, J.A. Haustein, S.S. Srinivasa, and T.~Asfour.
\newblock Nonprehensile whole arm rearrangement planning on physics manifolds.
\newblock In \emph{{IEEE} {ICRA}}, 2015.

\bibitem[La{V}alle(1998)]{lavalle98rrt}
S.M. La{V}alle.
\newblock {R}apidly-exploring {R}andom {T}rees: A new tool for path planning.
\newblock 1998.

\bibitem[Likhachev and Ferguson(2008)]{likhachev08lattice}
M.~Likhachev and D.~Ferguson.
\newblock Planning long dynamically-feasible maneuvers for autonomous vehicles.
\newblock In \emph{{RSS}}, 2008.

\bibitem[Likhachev et~al.(2004)Likhachev, Gordon, and Thrun]{likhachev04ara}
M.~Likhachev, G.~Gordon, and S.~Thrun.
\newblock {ARA}*: Anytime {A}* with provable bounds on sub-optimality.
\newblock In \emph{{NIPS}}, 2004.

\bibitem[Lynch(1997)]{lynch97locallycontrol}
K.~Lynch.
\newblock Locally controllable polygons by stable pushing.
\newblock In \emph{{IEEE} {ICRA}}, 1997.

\bibitem[Lynch and Mason(1995{\natexlab{a}})]{lynch95controllability}
K.~Lynch and M.T. Mason.
\newblock Controllability of pushing.
\newblock In \emph{{IEEE} {ICRA}}, 1995{\natexlab{a}}.

\bibitem[Lynch and Mason(1995{\natexlab{b}})]{lynch95stablepushing}
K.~Lynch and M.T. Mason.
\newblock Stable pushing: {M}echanics, controllability, and planning.
\newblock In \emph{{WAFR}}, 1995{\natexlab{b}}.

\bibitem[Mason(1986)]{mason86plan}
M.T. Mason.
\newblock Mechanics and planning of manipulator pushing operations.
\newblock \emph{{IJRR}}, 5\penalty0 (3):\penalty0 53--71, 1986.

\bibitem[Nieuwenhuisen et~al.(2005)Nieuwenhuisen, van~der Stappen, and
  Overmars]{nieuwenhuisen05discpush}
D.~Nieuwenhuisen, A.~Frank van~der Stappen, and M.H. Overmars.
\newblock Path planning for pushing a disk using compliance.
\newblock In \emph{{IEEE/RSJ} {IROS}}, 2005.

\bibitem[Nieuwenhuisen et~al.(2008)Nieuwenhuisen, Stappen., and
  Overmars]{nieuwenhuisen08namo}
D.~Nieuwenhuisen, A.~Stappen., and M.~Overmars.
\newblock An effective framework for path planning amidst movable obstacles.
\newblock In \emph{{WAFR}}, 2008.

\bibitem[Olson(2011)]{olson11tags}
E.~Olson.
\newblock {AprilTag}: A robust and flexible visual fiducial system.
\newblock In \emph{{IEEE} {ICRA}}, 2011.

\bibitem[Peshkin and Sanderson(1988)]{peshkin88slide}
M.A. Peshkin and A.C. Sanderson.
\newblock Planning robotic manipulation strategies for workpieces that slide.
\newblock \emph{{IJRA}}, 4\penalty0 (5):\penalty0 524--531, 1988.

\bibitem[Pohl(1970)]{pohl70weightedastar}
I.~Pohl.
\newblock Heuristic search viewed as path finding in a graph.
\newblock \emph{Artificial Intelligence}, 1\penalty0 (3):\penalty0 193--204,
  1970.

\bibitem[S\'{a}nchez and Latombe(2002)]{sanchez02collisionchecking}
G.~S\'{a}nchez and J-C. Latombe.
\newblock On delaying collision checking in {PRM} planning - application to
  multi-robot coordination.
\newblock \emph{{IJRR}}, 21\penalty0 (1):\penalty0 5--26, 2002.

\bibitem[Sekhavat et~al.(1998)Sekhavat, \v{S}vestka, Laumond, and
  Overmars]{sekhavat98multilevel}
S.~Sekhavat, P.~\v{S}vestka, J.-P. Laumond, and M.H. Overmars.
\newblock Multi-level path planning for nonholonomic robots using
  semi-holonomic subsystems.
\newblock \emph{{IJRR}}, 17\penalty0 (8):\penalty0 840--857, 1998.

\bibitem[Sim\'eon et~al.(2004)Sim\'eon, Laumond, Cort\'es, and
  Sahbani]{simeon04prm}
T.~Sim\'eon, J-P. Laumond, J.~Cort\'es, and A.~Sahbani.
\newblock Manipulation planning with probabilistic roadmaps.
\newblock \emph{{IJRR}}, 23\penalty0 (7--8):\penalty0 729--746, 2004.

\bibitem[Stilman. and Kuffner(2004)]{stilman04namo}
M.~Stilman. and J.~Kuffner.
\newblock Navigation among movable obstacles: Real-time reasoning in complex
  environments.
\newblock In \emph{{IEEE-RAS} Humanoids}, 2004.

\bibitem[Stilman et~al.(2007)Stilman, Schamburek, Kuffner, and
  Asfour]{stilman07namo}
M.~Stilman, J.~Schamburek, J.~Kuffner, and T.~Asfour.
\newblock Manipulation planning among movable obstacles.
\newblock In \emph{{IEEE} {ICRA}}, 2007.

\bibitem[van~den Berg et~al.(2008)van~den Berg, Stilman, Kuffner, Lin, and
  Manocha]{vandenburg08planning}
J.~van~den Berg, M.~Stilman, J.~Kuffner, M.~Lin, and D.~Manocha.
\newblock Path planning among movable obstacles: a probabilistically complete
  approach.
\newblock In \emph{{WAFR}}, 2008.

\bibitem[Webb and van~den Berg(2012)]{webb12kinorrt}
D.~J. Webb and J.~van~den Berg.
\newblock Kinodynamic rrt*: Optimal motion planning for systems with linear
  differential constraints.
\newblock \emph{{C}o{RR}}, 2012.

\bibitem[Zickler and Velosa(2009)]{zickler09physics}
S.~Zickler and M.~Velosa.
\newblock Efficient physics-based planning: sampling search via
  non-deterministic tactics and skills.
\newblock In \emph{{AAMAS}}, 2009.

\end{thebibliography}
